%% file: iclr2025_conference.tex
\documentclass{article} 
\usepackage{iclr2025_conference,times}

\input{math_commands.tex}

\usepackage{hyperref}
\usepackage{url}
\usepackage{graphicx}
\usepackage{wrapfig}
\usepackage{hyperref}
\usepackage{subcaption}
\usepackage{listings}
\usepackage{cleveref}
\usepackage{placeins}

\title{\centering Reasoning Effort and Problem Complexity:\\A Scaling Analysis in LLMs}

\iclrfinalcopy

\author{%
 {Benjamin Estermann} \\
 ETH Zürich\\
 Switzerland \\
 \texttt{estermann@ethz.ch} \\
 \And
 {Roger Wattenhofer} \\
 ETH Zürich\\
 Switzerland \\
 \texttt{wattenhofer@ethz.ch} \\
}

\begin{document}

\maketitle

\begin{abstract}
Large Language Models (LLMs) have demonstrated remarkable text generation capabilities, and recent advances in training paradigms have led to breakthroughs in their reasoning performance.
In this work, we investigate how the reasoning effort of such models scales with problem complexity.
We use the infinitely scalable Tents puzzle, which has a known linear-time solution, to analyze this scaling behavior.
Our results show that reasoning effort scales with problem size, but only up to a critical problem complexity.
Beyond this threshold, the reasoning effort does not continue to increase, and may even decrease.
This observation highlights a critical limitation in the logical coherence of current LLMs as problem complexity increases, and underscores the need for strategies to improve reasoning scalability.
Furthermore, our results reveal significant performance differences between current state-of-the-art reasoning models when faced with increasingly complex logical puzzles.
\end{abstract}

\section{Introduction}
Large language models (LLMs) have demonstrated remarkable abilities in a wide range of natural language tasks, from text generation to complex problem-solving.
Recent advances, particularly with models trained for enhanced reasoning, have pushed the boundaries of what machines can achieve in tasks requiring logical inference and deduction.
\begin{wrapfigure}{r}{0.35\textwidth}
    \vspace{-13pt}
  \begin{center}
    \includegraphics[width=0.33\textwidth]{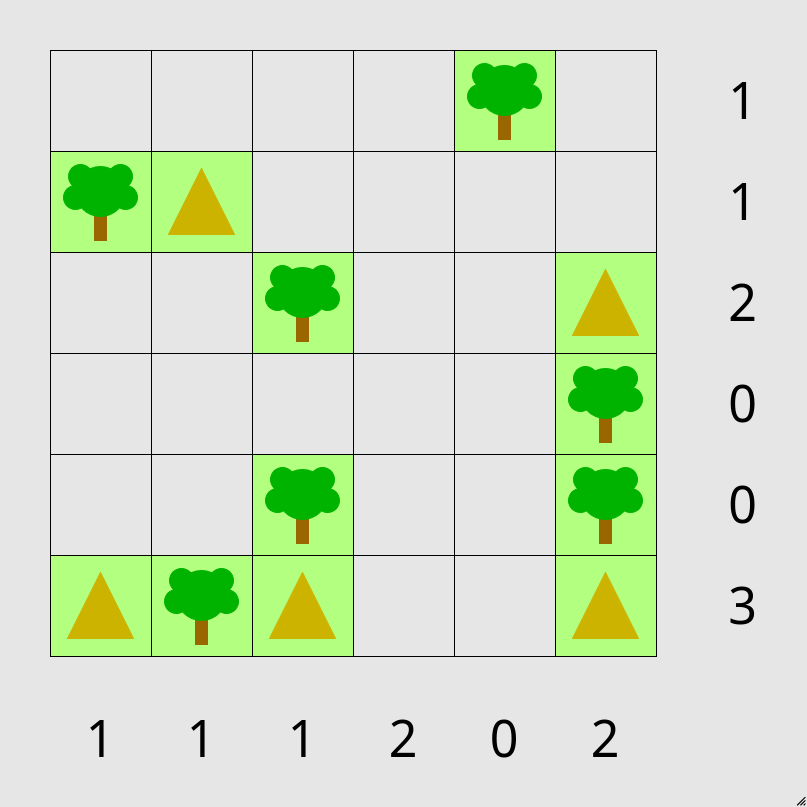}
  \end{center}
  \caption{An example instance of a partially solved 6 by 6 tents puzzle. Tents need to be placed next to trees, away from other tents and fulfilling the row and column constraints.}
  \vspace{-23pt}
  \label{fig:tents-example}
\end{wrapfigure}
A critical factor in the success of these advanced models is the ability to leverage increased computational resources at test time, allowing them to explore more intricate solution spaces.
This capability raises a fundamental question: \textit{how does the "reasoning effort" of these models scale as the complexity of the problem increases?}

Understanding this scaling relationship is crucial for several reasons.
First, it sheds light on the fundamental nature of reasoning within LLMs, moving beyond simply measuring accuracy on isolated tasks.
By examining how the computational demands, reflected in token usage, evolve with problem difficulty, we can gain insights into the efficiency and potential bottlenecks of current LLM architectures.
Second, characterizing this scaling behavior is essential for designing more effective and resource-efficient reasoning models in the future.

In this work, we address this question by investigating the scaling of reasoning effort in LLMs using a specific, infinitely scalable logic puzzle: the Tents puzzle\footnote{The puzzle is available to play in the browser at \url{https://www.chiark.greenend.org.uk/~sgtatham/puzzles/js/tents.html}} (see Figure~\ref{fig:tents-example}).
This puzzle offers a controlled environment for studying algorithmic reasoning, as its problem size can be systematically increased, and it possesses a known linear-time solution. 
Our analysis focuses on how the number of tokens used by state-of-the-art reasoning LLMs changes as the puzzle grid size grows. 
In addition to reasoning effort, we also evaluate the success rate across different puzzle sizes to provide a comprehensive view of their performance.

\section{Related Work}
The exploration of reasoning abilities in large language models (LLMs) is a rapidly evolving field with significant implications for artificial intelligence. 
Several benchmarks have been developed to evaluate the reasoning capabilities of LLMs across various domains. These benchmarks provide standardized tasks and evaluation metrics to assess and compare different models. 
Notable benchmarks include GSM8K~\citep{cobbe2021training}, ARC-AGI~\citep{chollet2019measure}, GPQA~\citep{rein2023gpqa}, MMLU~\citep{hendrycks2020measuring}, SWE-bench~\citep{jimenez2023swe} and NPhard-eval~\citep{fan2023nphardeval}.
These benchmarks cover topics from mathematics to commonsense reasoning and coding.
More recently, also math competitions such as AIME2024~\citep{AoPS_AIME_2024} have been used to evaluate the newest models.
\citet{estermann2024puzzles} have introduced PUZZLES, a benchmark focusing on algorithmic and logical reasoning for reinforcement learning.
While PUZZLES does not focus on LLMs, except for a short ablation in the appendix, we argue that the scalability provided by the underlying puzzles is an ideal testbed for testing extrapolative reasoning capabilities in LLMs.

The reasoning capabilities of traditional LLMs without specific prompting strategies are quite limited \citep{huang2022towards}.
Using prompt techniques such as chain-of-thought~\citep{wei2022chain}, least-to-most~\citep{zhou2022least} and tree-of-thought~\citep{yao2023tree}, the reasoning capabilities of traditional LLMs can be greatly improved. \citet{lee2024reasoning} have introduced the Language of Thought Hypothesis, based on the idea that human reasoning is rooted in language. Lee et al. propose to see the reasoning capabilities through three different properties: Logical coherence, compositionality and productivity. In this work we will mostly focus on the aspect of algorithmic reasoning, which falls into logical coherence. Specifically, we analyze the limits of logical coherence.

With the release of OpenAI's o1 model, it became apparent that new training strategies based on reinforcement learning are able to boost the reasoning performance even further.
Since o1, there now exist a number of different models capable of enhanced reasoning~\citep{guo2025deepseek,deepmind_gemini_flash_thinking,qwq-32b-preview,openai_o3_mini}.
Key to the success of these models is the scaling of test-time compute. 
Instead of directly producing an answer, or thinking for a few steps using chain-of-thought, the models are now trained to think using several thousands of tokens before coming up with an answer.



\section{Methods}

\subsection{The Tents Puzzle Problem}

In this work, we employ the Tents puzzle, a logic problem that is both infinitely scalable and solvable in linear time\footnote{See a description of the algorithm of the solver as part of the PUZZLES benchmark here: \url{https://github.com/ETH-DISCO/rlp/blob/main/puzzles/tents.c\#L206C3-L206C67}}, making it an ideal testbed for studying algorithmic reasoning in LLMs.
The Tents puzzle, popularized by Simon Tatham's Portable Puzzle Collection~\citep{simon_tatham_portable_puzzle_collection}, requires deductive reasoning to solve.  The puzzle is played on a rectangular grid, where some cells are pre-filled with trees. The objective is to place tents in the remaining empty cells according to the following rules:
\begin{itemize}
    \item no two tents are adjacent, even diagonally
    \item there are exactly as many tents as trees and the number of tents in each row and column matches the numbers around the edge of the grid
    \item it is possible to match all tents to trees so that each tent is orthogonally adjacent to its own tree (a tree may also be adjacent to other tents). 
\end{itemize}
An example instance of the Tents puzzle is visualized in Figure~\ref{fig:tents-example} in the Introduction.
The scalability of the puzzle is achieved by varying the grid dimensions, allowing for systematic exploration of problem complexity.
Where not otherwise specified, we used the "easy" difficulty preset available in the Tents puzzle generator, with "tricky" being evaluated in \Cref{app:easy-vs-tricky}.
Crucially, the Tents puzzle is designed to test extrapolative reasoning as puzzle instances, especially larger ones, are unlikely to be present in the training data of LLMs.
We utilized a text-based interface for the Tents puzzle, extending the code base provided by \citet{estermann2024puzzles} to generate puzzle instances and represent the puzzle state in a format suitable for LLMs.

Models were presented with the same prompt (detailed in Appendix~\ref{sec:appendix-prompt}) for all puzzle sizes and models tested.
The prompt included the puzzle rules and the initial puzzle state in a textual format.
Models were tasked with directly outputting the solved puzzle grid in JSON format.
This one-shot approach contrasts with interactive or cursor-based approaches previously used in \citep{estermann2024puzzles}, isolating the reasoning process from potential planning or action selection complexities.

\subsection{Evaluation Criteria}

Our evaluation focuses on two key metrics: \textit{success rate} and \textit{reasoning effort}.  Success is assessed as a binary measure: whether the LLM successfully outputs a valid solution to the Tents puzzle instance, adhering to all puzzle rules and constraints.  We quantify problem complexity by the \textit{problem size}, defined as the product of the grid dimensions (rows $\times$ columns).  To analyze the scaling of reasoning effort, we measure the \textit{total number of tokens} generated by the LLMs to produce the final answer, including all thinking tokens.
We hypothesize a linear scaling relationship between problem size and reasoning effort, and evaluate this hypothesis by fitting a linear model to the observed token counts as a function of problem size.
The goodness of fit is quantified using the $R^2$ metric, where scores closer to 1 indicate that a larger proportion of the variance in reasoning effort is explained by a linear relationship with problem size.
Furthermore, we analyze the percentage of correctly solved puzzles across different problem sizes to assess the performance limits of each model.

\subsection{Considered Models}

We evaluated the reasoning performance of the following large language models known for their enhanced reasoning capabilities: Gemini 2.0 Flash Thinking~\citep{deepmind_gemini_flash_thinking}, OpenAI o3-mini~\citep{openai_o3_mini}, DeepSeek R1~\cite{guo2025deepseek}, and Qwen/QwQ-32B-Preview~\cite{qwq-32b-preview}.

\section{Results}

\begin{figure}[hb]
    \centering
    \begin{subfigure}[b]{0.49\textwidth}
        \centering
        \includegraphics[width=0.96\linewidth]{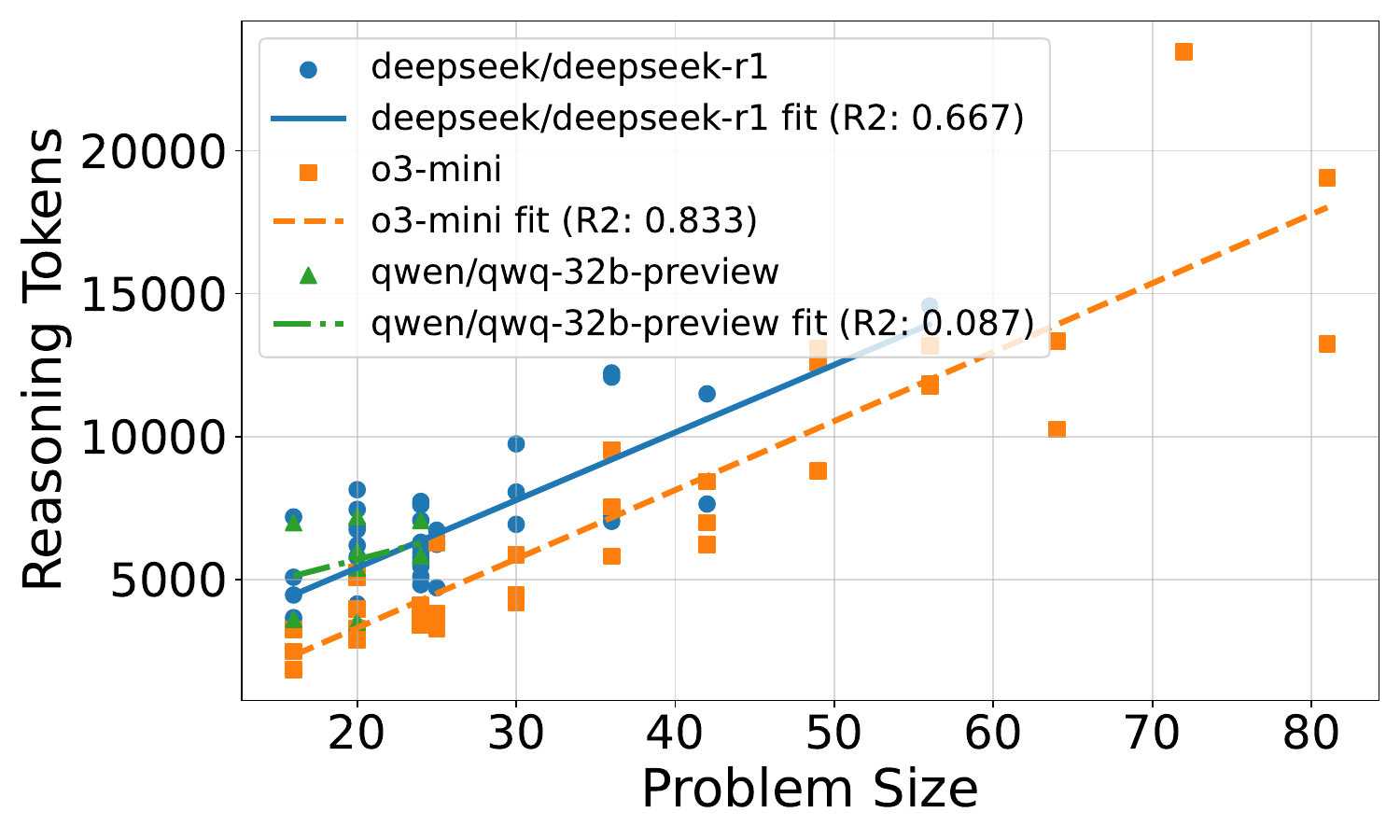}
        \caption{}
        \label{fig:reasoning-effort}
    \end{subfigure}
    \hfill
    \begin{subfigure}[b]{0.49\textwidth}
        \centering
        \includegraphics[width=0.96\linewidth]{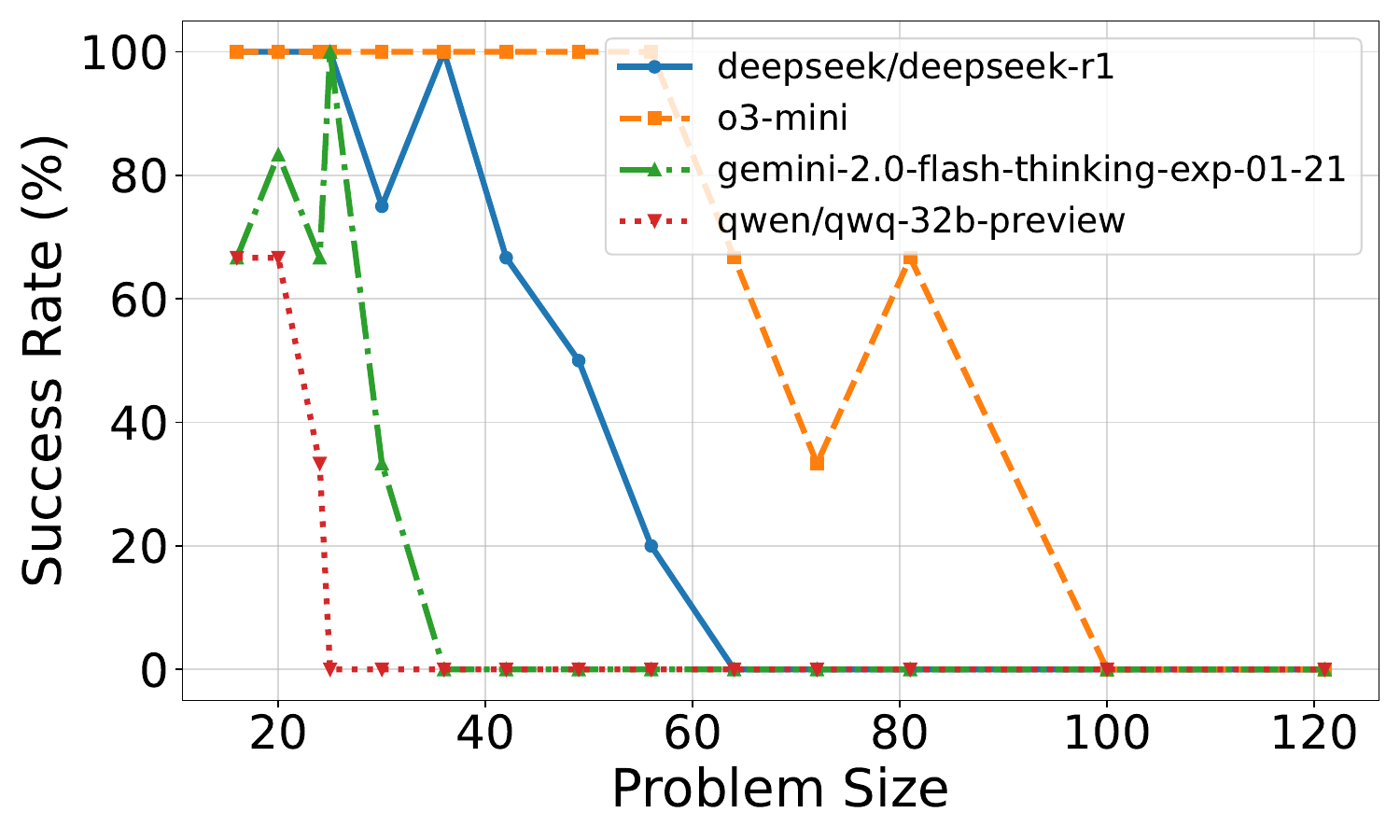}
        \caption{}
        \label{fig:solved-percentage}
    \end{subfigure}
    \caption{
    (a) Reasoning effort in number of reasoning tokens versus problem size for DeepSeek R1, o3-mini, and Qwen/QwQ-32B-Preview. Successful attempts only. Linear fits are added for each model. Gemini 2.0 Flash Thinking is excluded due to unknown number of thinking tokens.\\
    (b) Solved percentage versus problem size for all models. No model solved problems larger than size 100. o3-mini achieves the highest success rate, followed by DeepSeek R1 and Gemini 2.0 Flash Thinking. Qwen/QwQ-32B-Preview struggles with problem instances larger than size 20.}
    \label{fig:overall-effort-solved}
\end{figure}

The relationship between reasoning effort and problem size reveals interesting scaling behaviors across the evaluated models. 
Figure~\ref{fig:reasoning-effort} illustrates the scaling of reasoning effort, measured by the number of reasoning tokens, as the problem size increases for successfully solved puzzles.
For DeepSeek R1 and o3-mini, we observe a roughly linear increase in reasoning effort with problem size.
Notably, the slopes of the linear fits for R1 and o3-mini are very similar, suggesting comparable scaling behavior in reasoning effort for these models, although DeepSeek R1 consistently uses more tokens than o3-mini across problem sizes.
Qwen/QwQ-32B-Preview shows a weaker linear correlation, likely due to the limited number of larger puzzles it could solve successfully.

The problem-solving capability of the models, shown in Figure~\ref{fig:solved-percentage}, reveals performance limits as problem size increases.  None of the models solved puzzles with a problem size exceeding 100.  o3-mini demonstrates the highest overall solvability, managing to solve the largest problem instances, followed by DeepSeek R1 and Gemini 2.0 Flash Thinking. Qwen/QwQ-32B-Preview's performance significantly degrades with increasing problem size, struggling to solve instances larger than 25.

\begin{figure}[ht]
    \centering
    \begin{subfigure}[b]{0.49\textwidth}
        \centering
        \includegraphics[width=0.96\linewidth]{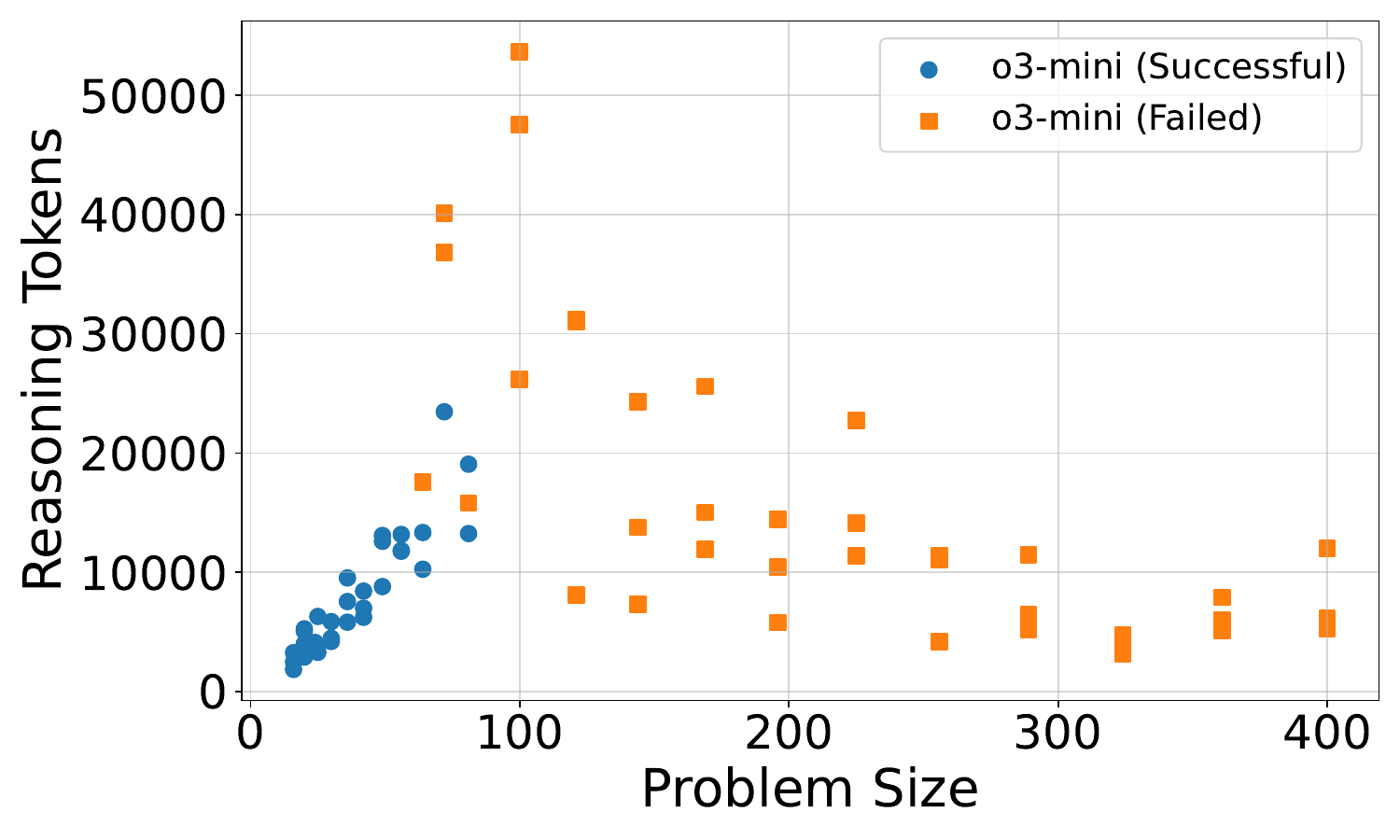}
        \caption{}
        \label{fig:o3-scaling}
    \end{subfigure}
    \hfill
    \begin{subfigure}[b]{0.49\textwidth}
        \centering
        \includegraphics[width=0.96\linewidth]{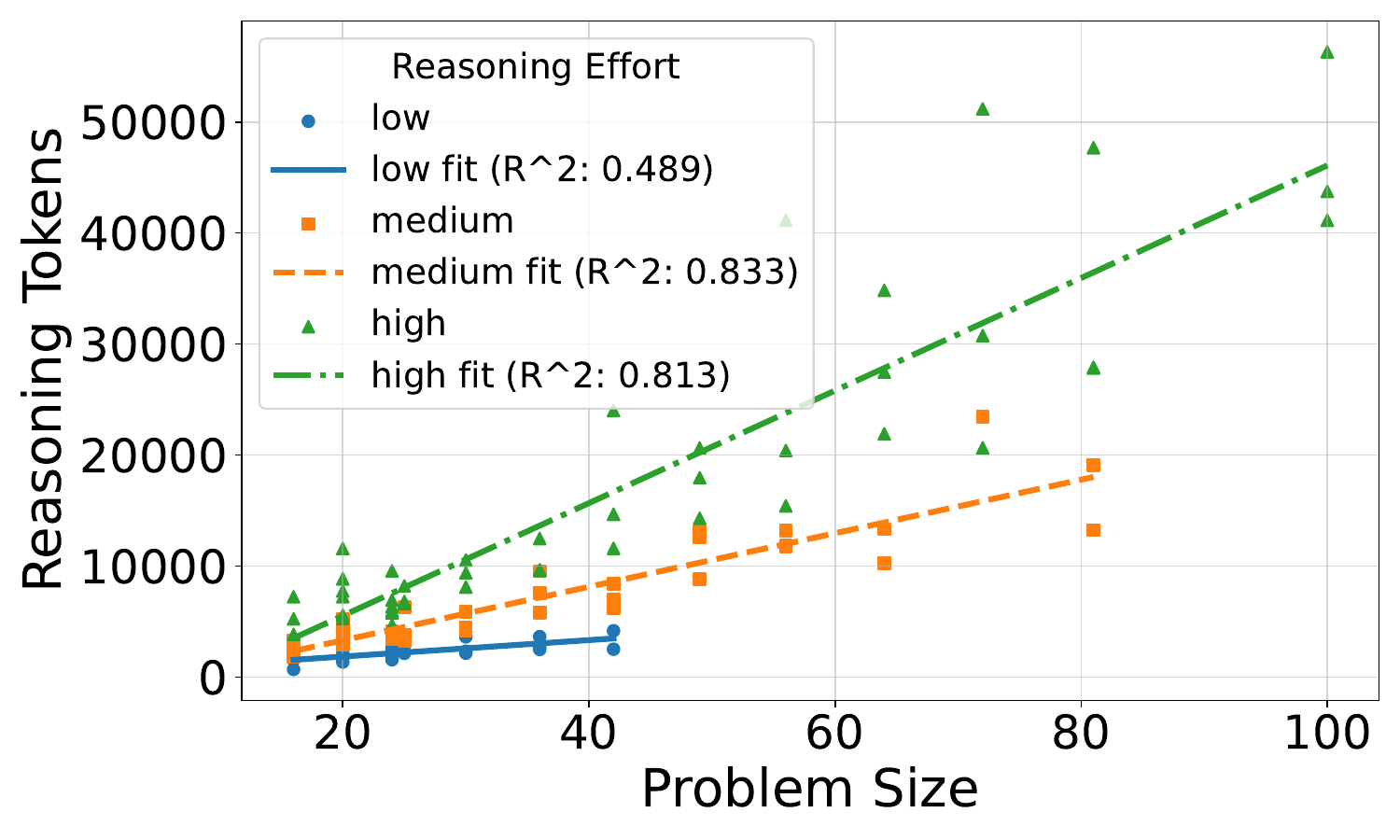}
        \caption{}
        \label{fig:o3-reasoning-effort}
    \end{subfigure}
    \caption{
    (a) Reasoning effort in number of reasoning tokens versus problem size for o3-mini. A peak in reasoning effort is observed around problem size 100, followed by a decline for larger problem sizes.
    (b) Reasoning effort in number of reasoning tokens versus problem size for o3-mini, categorized by low, medium, and high reasoning effort strategies. Steeper slopes are observed for higher reasoning effort strategies. High reasoning effort enables solving larger instances but also increases token usage for smaller, already solvable problems.}
    \label{fig:o3-mini-effort-solved}
\end{figure}

A more detailed analysis of o3-mini's reasoning effort (Figure~\ref{fig:o3-scaling}) reveals a non-monotonic trend. While generally increasing with problem size initially, reasoning effort peaks around a problem size of 100. Beyond this point, the reasoning effort decreases, suggesting a potential "frustration" effect where increased complexity no longer leads to proportionally increased reasoning in the model. The same behavior could not be observed for other models, see \Cref{app:reasoning-all-models}.
It would be interesting to see the effect of recent works trying to optimize reasoning length would have on these results \citep{luo2025o1}.

Figure~\ref{fig:o3-reasoning-effort} further explores o3-mini's behavior by categorizing reasoning effort into low, medium, and high strategies.  The steepness of the scaling slope increases with reasoning effort, indicating that higher effort strategies lead to a more pronounced increase in token usage as problem size grows. While high reasoning effort enables solving larger puzzles (up to 10x10), it also results in a higher token count even for smaller problems that were already solvable with lower effort strategies. This suggests a trade-off where increased reasoning effort can extend the solvable problem range but may also introduce inefficiencies for simpler instances.

\section{Conclusion}
This study examined how reasoning effort scales in LLMs using the Tents puzzle.
We found that reasoning effort generally scales linearly with problem size for solvable instances.
Model performance varied, with o3-mini and DeepSeek R1 showing better performance than Qwen/QwQ-32B-Preview and Gemini 2.0 Flash Thinking.
These results suggest that while LLMs can adapt reasoning effort to problem complexity, their logical coherence has limits, especially for larger problems.
Future work should extend this analysis to a wider variety of puzzles contained in the PUZZLES benchmark to include puzzles with different algorithmic complexity.
These insights could lead the way to find strategies to improve reasoning scalability and efficiency, potentially by optimizing reasoning length or refining prompting techniques.
Understanding these limitations is crucial for advancing LLMs in complex problem-solving.


\bibliography{iclr2025_conference}
\bibliographystyle{iclr2025_conference}

\clearpage
\appendix
\section{Appendix}

\subsection{Full Prompt}
\label{sec:appendix-prompt}

The full prompt used in the experiments is the following, on the example of a 4x4 puzzle:
\begin{lstlisting}[breaklines=true, breakindent=0pt]
You are a logic puzzle expert. You will be given a logic puzzle to solve. Here is a description of the puzzle:
You have a grid of squares, some of which contain trees. Your aim is to place tents in some of the remaining squares, in such a way that the following conditions are met:
There are exactly as many tents as trees.
The tents and trees can be matched up in such a way that each tent is directly adjacent (horizontally or vertically, but not diagonally) to its own tree. However, a tent may be adjacent to other trees as well as its own.
No two tents are adjacent horizontally, vertically or diagonally.
The number of tents in each row, and in each column, matches the numbers given in the row or column constraints.
Grass indicates that there cannot be a tent in that position.
You receive an array representation of the puzzle state as a grid. Your task is to solve the puzzle by filling out the grid with the correct values. You need to solve the puzzle on your own, you cannot use any external resources or run any code. Once you have solved the puzzle, tell me the final answer without explanation. Return the final answer as a JSON array of arrays.
Here is the current state of the puzzle as a string of the internal state representation:
A 0 represents an empty cell, a 1 represents a tree, a 2 represents a tent, and a 3 represents a grass patch.
Tents puzzle state:
Current grid:
[[0 0 1 0]
 [0 1 0 0]
 [1 0 0 0]
 [0 0 0 0]]
The column constraints are the following:
[1 1 0 1]
The row constraints are the following:
[2 0 0 1]
\end{lstlisting}

\subsection{Additional Figures}
\FloatBarrier
\subsubsection{Easy vs. Tricky Puzzles}
    \label{app:easy-vs-tricky}
    
    \begin{figure}[hb]
        \centering
        \begin{subfigure}[b]{0.49\textwidth}
            \centering
            \includegraphics[width=0.96\linewidth]{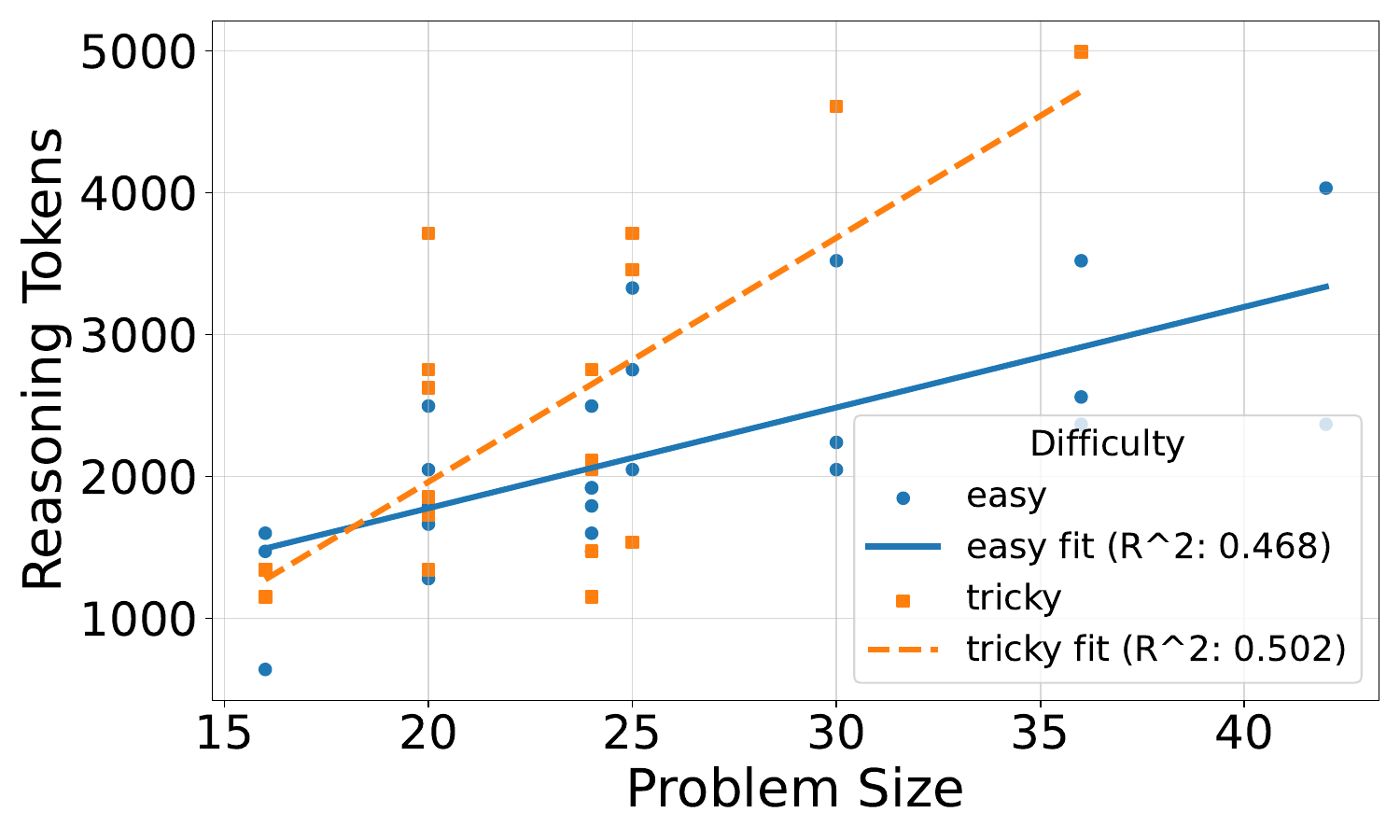}
            \caption{}
        \end{subfigure}
        \hfill
        \begin{subfigure}[b]{0.49\textwidth}
            \centering
            \includegraphics[width=0.96\linewidth]{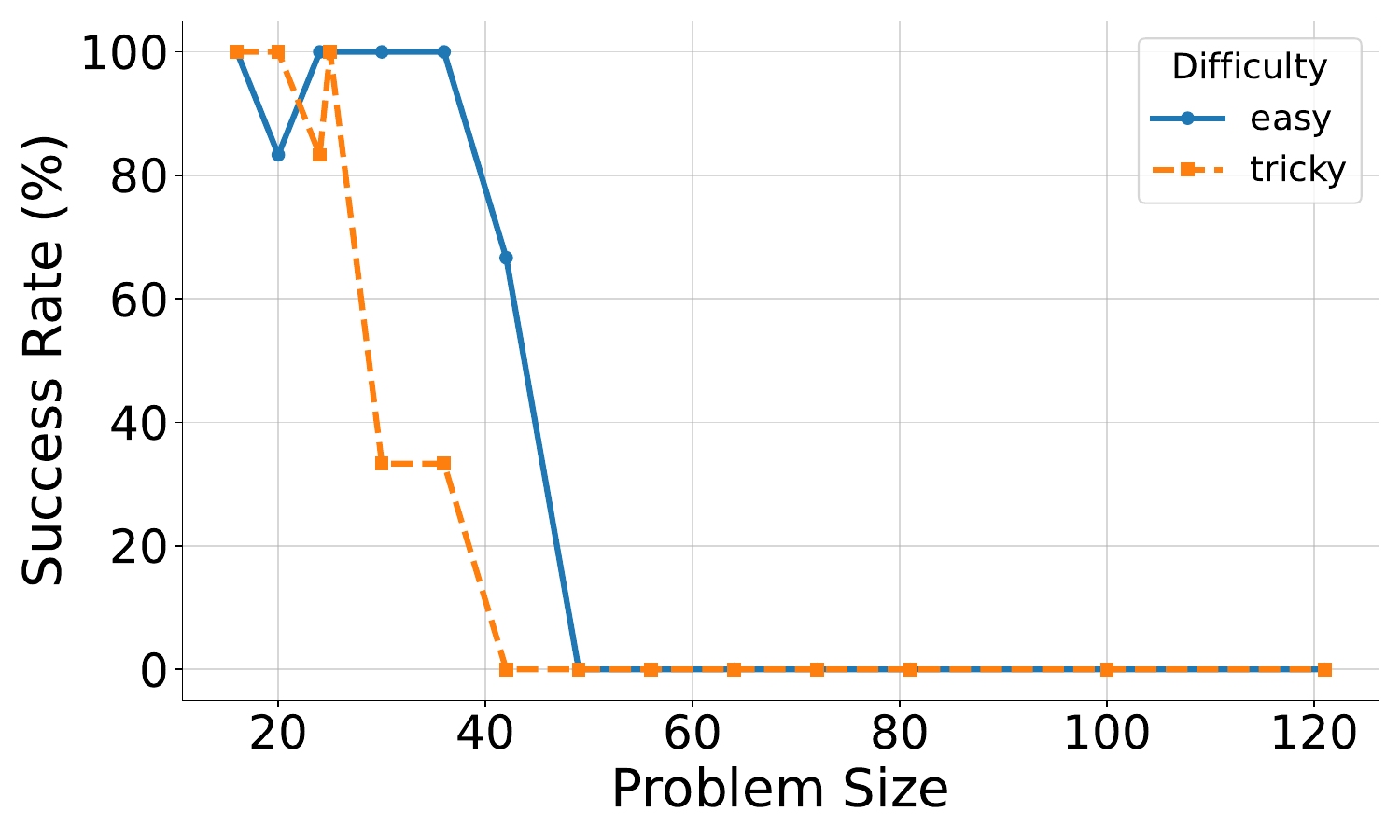}
            \caption{}
        \end{subfigure}
        \caption{
        (a) Reasoning effort in number of reasoning tokens versus problem size for o3-mini with reasoning effort \textbf{low}. Successful tries only. Linear fits are added for each model.
        (b) Solved percentage versus problem size for o3-mini with reasoning effort \textbf{low}.}
    \end{figure}
    \begin{figure}[hb]
        \centering
        \begin{subfigure}[b]{0.49\textwidth}
            \centering
            \includegraphics[width=0.96\linewidth]{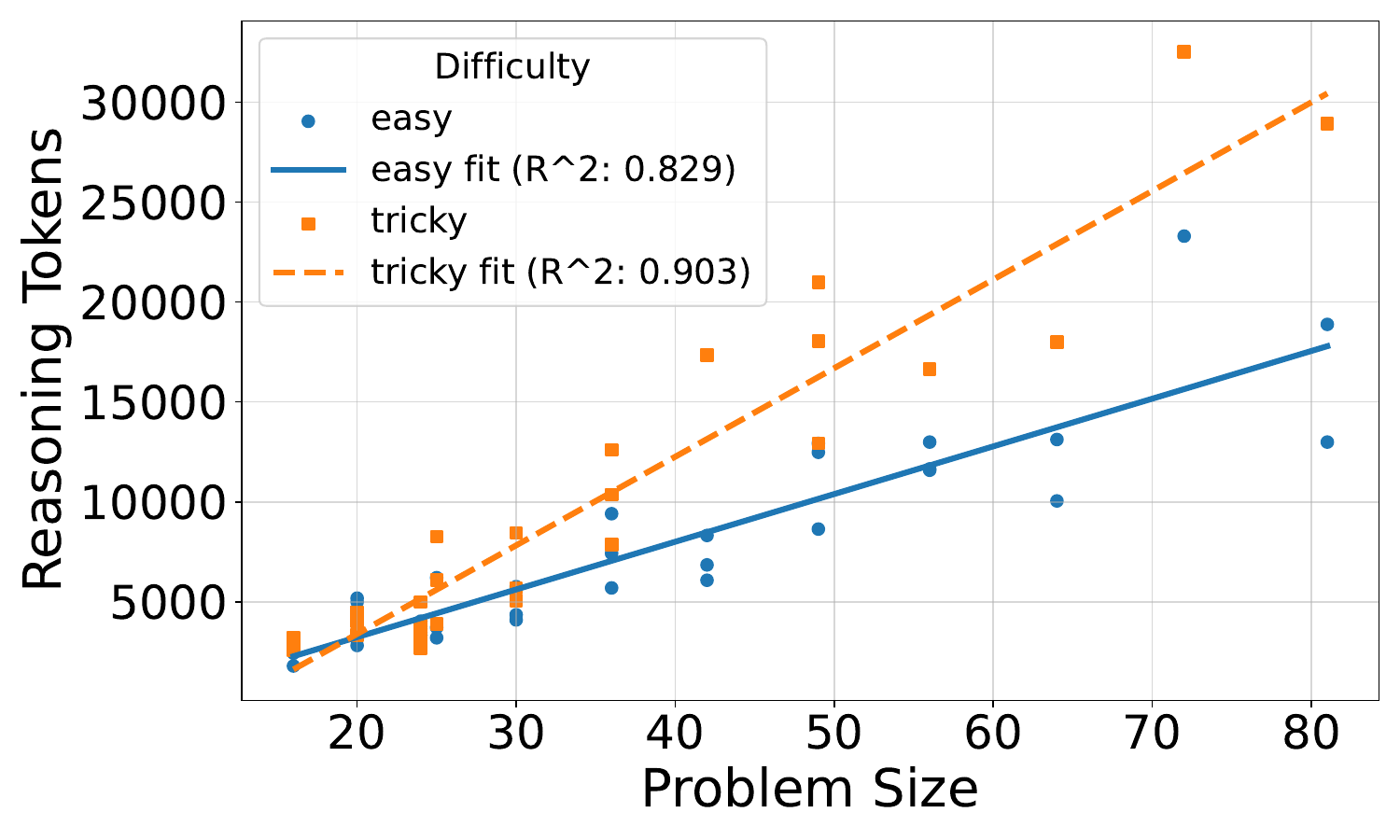}
            \caption{}
        \end{subfigure}
        \hfill
        \begin{subfigure}[b]{0.49\textwidth}
            \centering
            \includegraphics[width=0.96\linewidth]{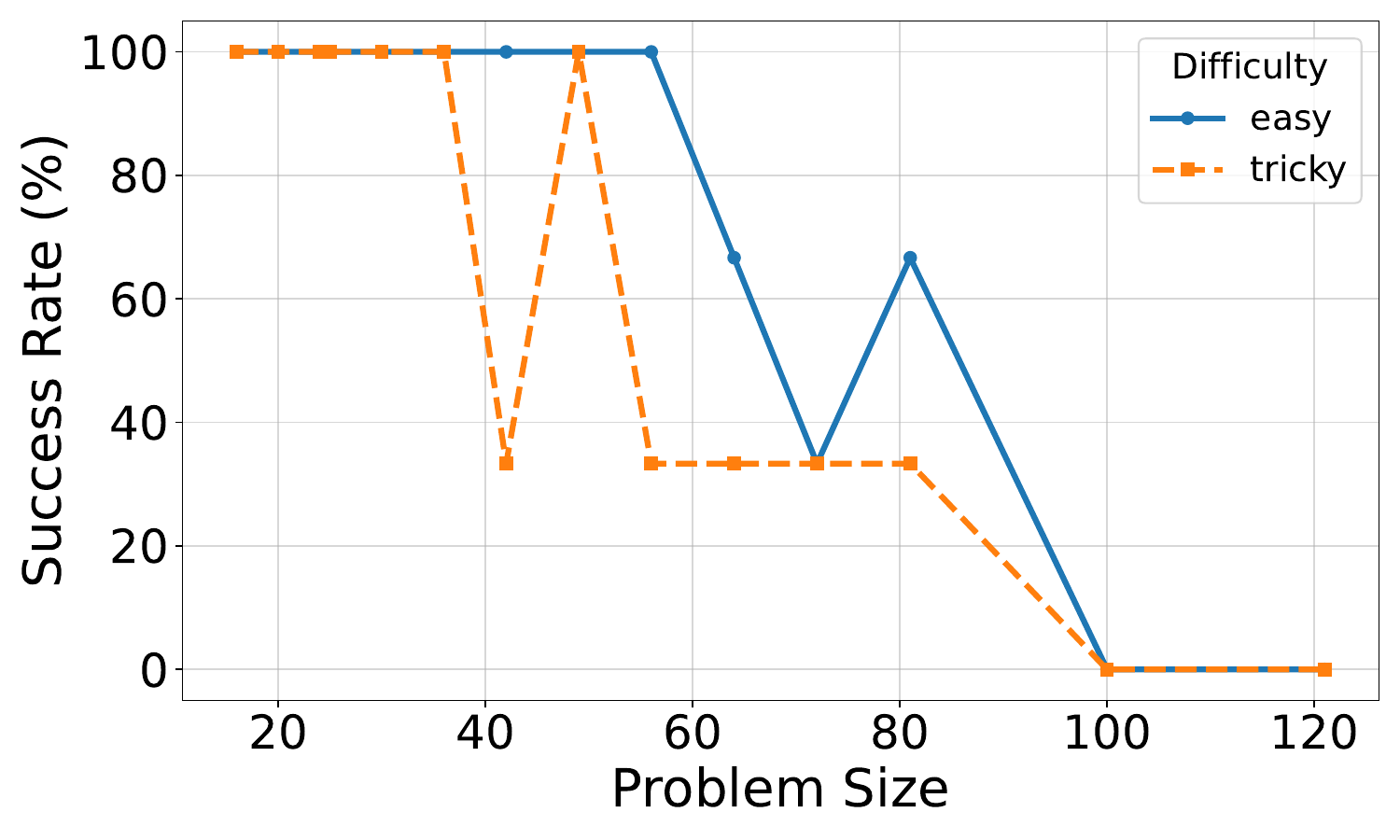}
            \caption{}
        \end{subfigure}
        \caption{
        (a) Reasoning effort in number of reasoning tokens versus problem size for o3-mini with reasoning effort \textbf{medium}. Successful tries only. Linear fits are added for each model.
        (b) Solved percentage versus problem size for o3-mini with reasoning effort \textbf{medium}.}
    \end{figure}
    \begin{figure}[hb]
        \centering
        \begin{subfigure}[b]{0.49\textwidth}
            \centering
            \includegraphics[width=0.96\linewidth]{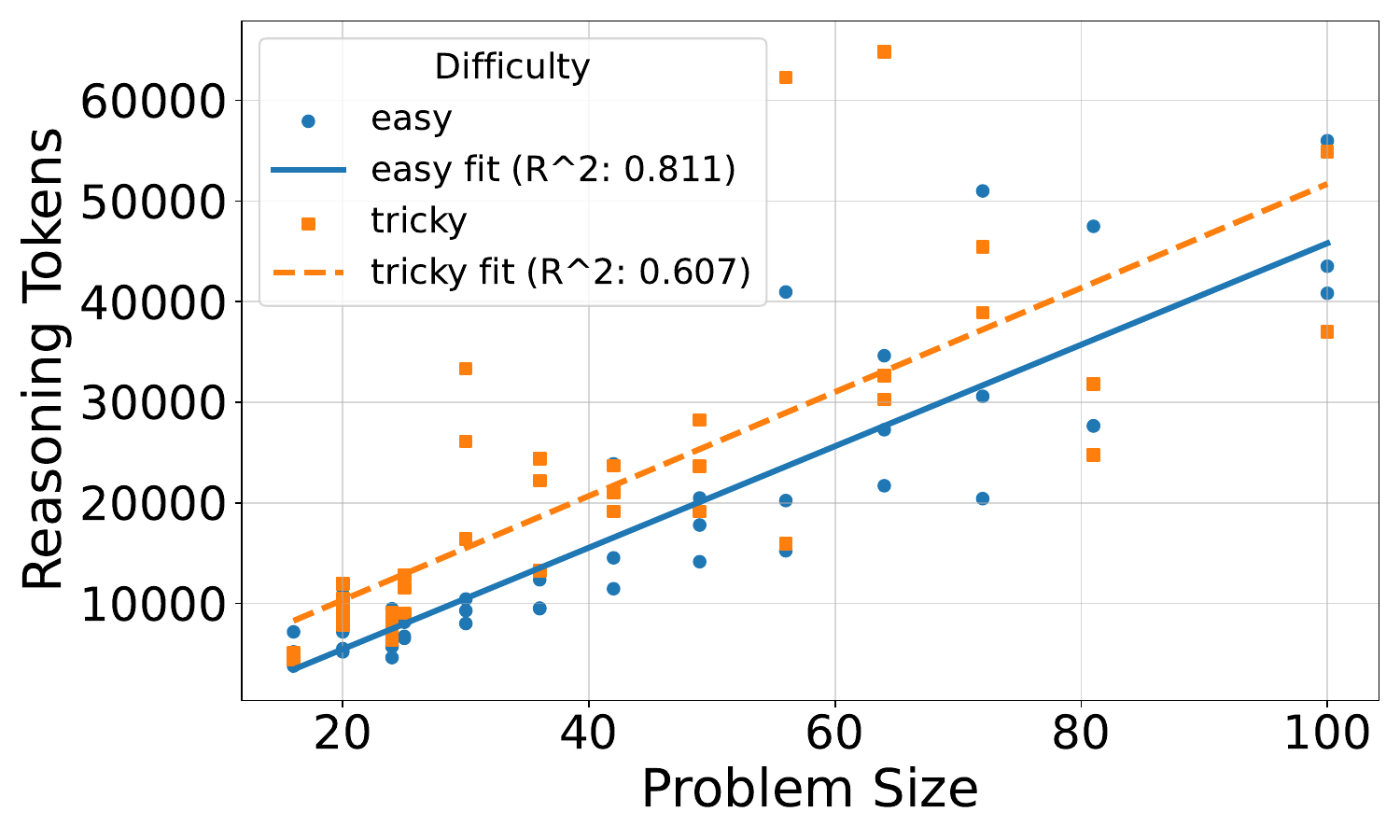}
            \caption{}
        \end{subfigure}
        \hfill
        \begin{subfigure}[b]{0.49\textwidth}
            \centering
            \includegraphics[width=0.96\linewidth]{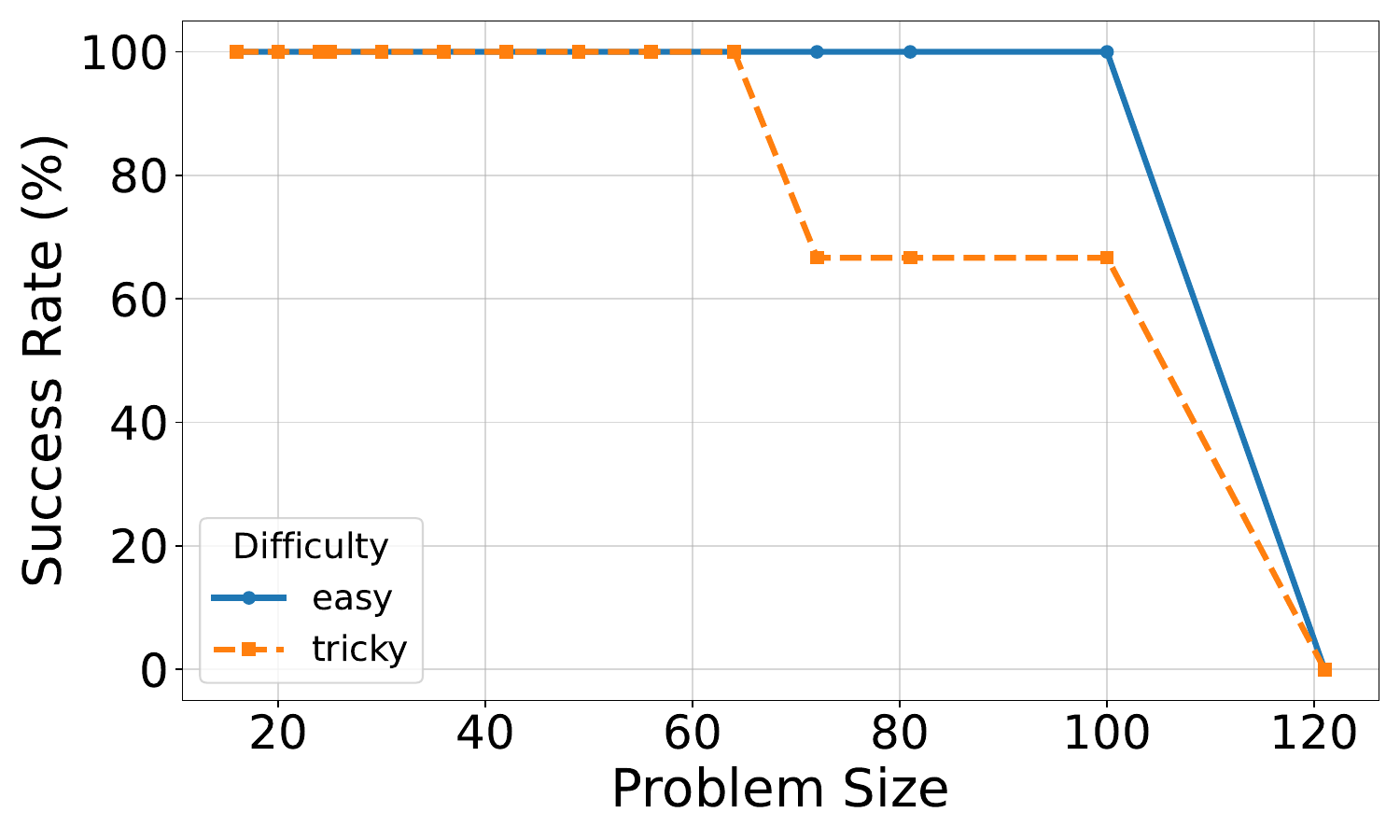}
            \caption{}
        \end{subfigure}
        \caption{
        (a) Reasoning effort in number of reasoning tokens versus problem size for o3-mini with reasoning effort \textbf{high}. Successful tries only. Linear fits are added for each model.
        (b) Solved percentage versus problem size for o3-mini with reasoning effort \textbf{high}.}
    \end{figure}
    \FloatBarrier

\subsubsection{Reasoning Effort for All Models}
\label{app:reasoning-all-models}
\begin{figure}[h]
    \centering
    \includegraphics[width=0.6\linewidth]{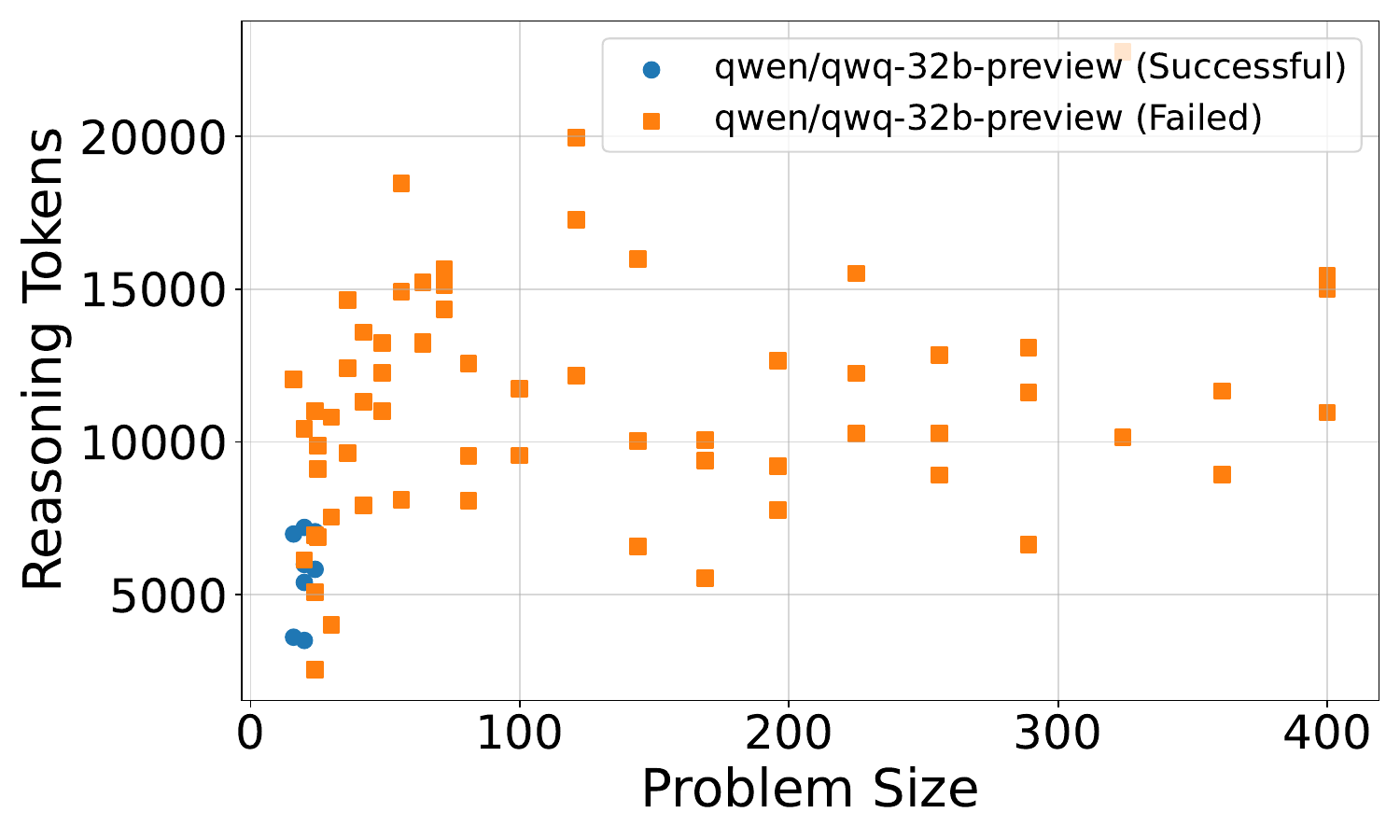}
    \caption{Reasoning effort in tokens for Qwen QwQ.}
\end{figure}
\begin{figure}[h]
    \centering
    \includegraphics[width=0.6\linewidth]{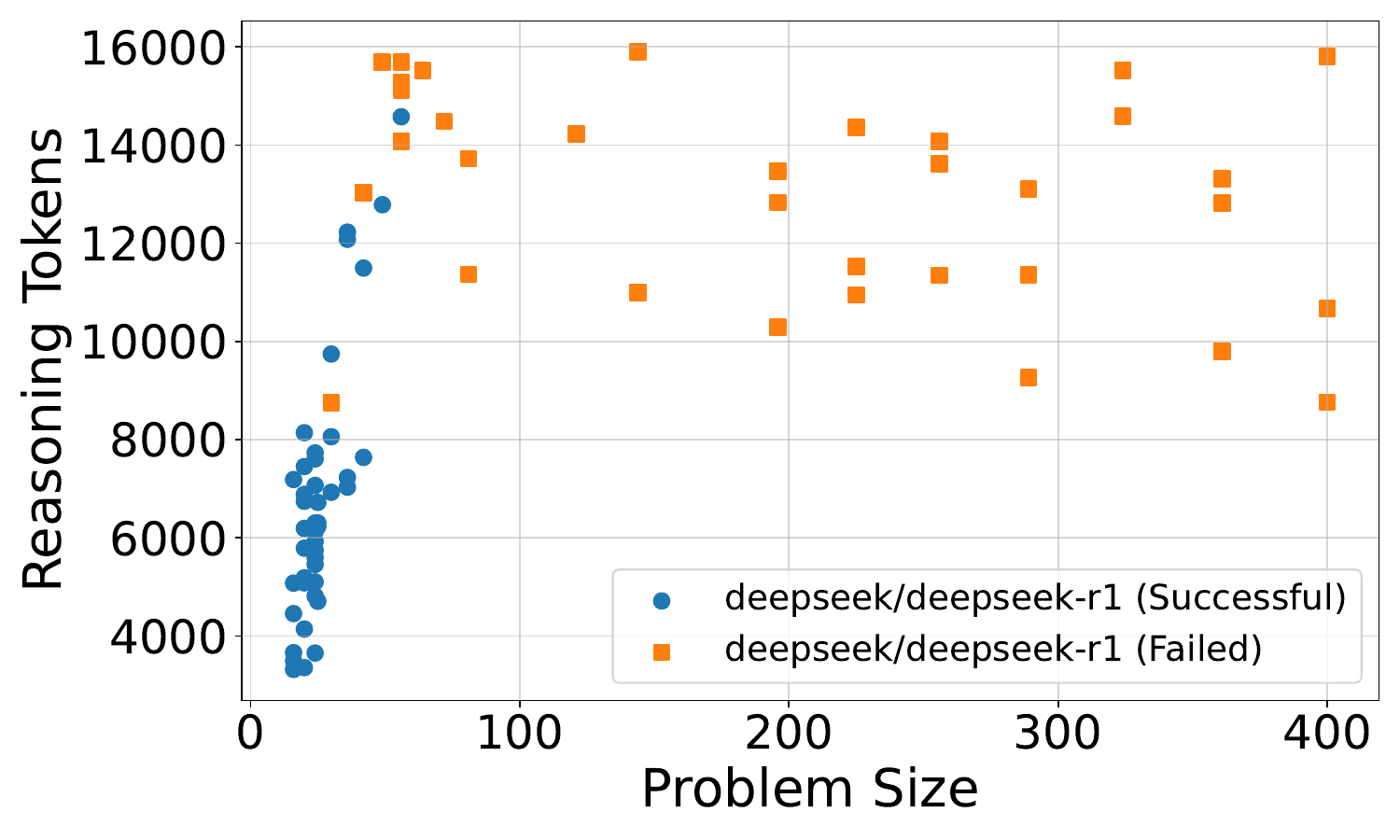}
    \caption{Reasoning effort in tokens for Deepseek R1.}
\end{figure}
\begin{figure}[h]
    \centering
    \includegraphics[width=0.6\linewidth]{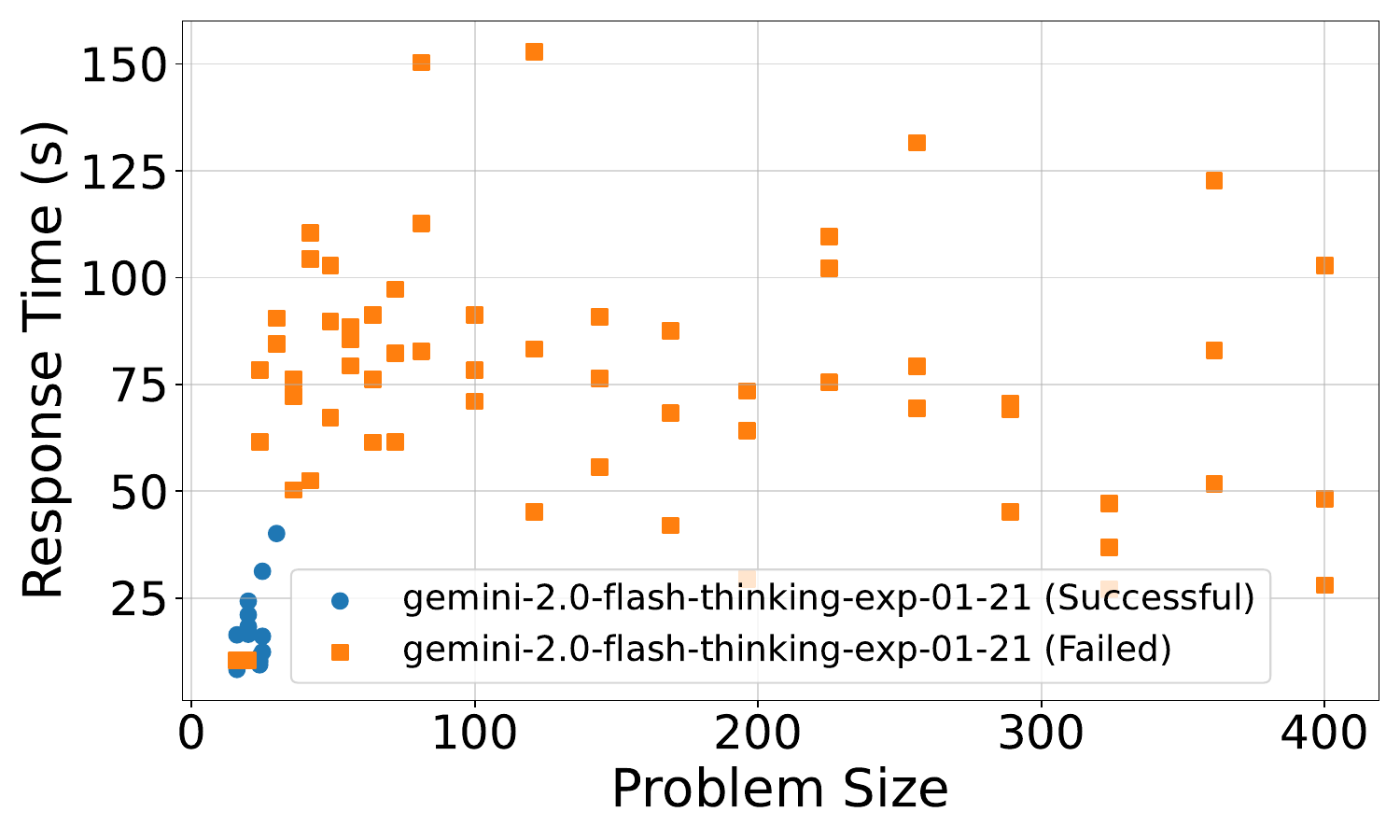}
    \caption{Reasoning effort quantified by response time for Gemini-2.0-flash-thinking.}
\end{figure}
\FloatBarrier

\subsection{Cost}
Total cost of these experiments was around 80 USD in API credits.

\end{document}

%% file: math_commands.tex
\usepackage{amsmath,amsfonts,bm}

\def\eqref#1{equation~\ref{#1}}

\def\1{\bm{1}}

\DeclareMathAlphabet{\mathsfit}{\encodingdefault}{\sfdefault}{m}{sl}
\SetMathAlphabet{\mathsfit}{bold}{\encodingdefault}{\sfdefault}{bx}{n}

